\documentclass[lettersize,journal]{IEEEtran}
\usepackage{amsmath,amsfonts}
\usepackage{algorithmic}
\usepackage{array}
\usepackage[caption=false,font=normalsize,labelfont=sf,textfont=sf]{subfig}
\usepackage{textcomp}
\usepackage{stfloats}
\usepackage{url}
\usepackage{verbatim}
\usepackage{graphicx}
\usepackage{booktabs}
\usepackage{times}
\usepackage{epsfig}
\usepackage{graphicx}
\usepackage{amsmath}
\usepackage{amssymb}
\usepackage[ruled]{algorithm2e}
\usepackage{multirow}
\usepackage{indentfirst}
\usepackage{graphicx}
\usepackage{float}
\usepackage{dirtree}
\usepackage{amsmath,amssymb}
\usepackage{setspace}
\usepackage{mathtools}
\usepackage{float}
\usepackage[normalem]{ulem}
\usepackage{array}
\usepackage{epstopdf}
\usepackage{amsmath}
\usepackage{times}
\usepackage{epsfig}
\usepackage{graphicx}
\usepackage{amsmath}
\usepackage{amssymb}
\usepackage{caption}
\usepackage{color}
\usepackage{subfig}
\usepackage{lipsum}
\usepackage{booktabs}
\usepackage{bm}
\usepackage{multicol}
\usepackage{multirow}
\usepackage{bbding}
\usepackage{array}
\usepackage{diagbox}
\usepackage{makecell}

\usepackage{cite}

\begin{document}

\title{3Deformer: A Common Framework for Image-Guided Mesh Deformation}

\author{Hao Su, Xuefeng Liu, Jianwei Niu$^*$, Ji Wan, Xinghao Wu}

\markboth{}{ }
\twocolumn[{%
\renewcommand\twocolumn[1][]{#1}%
\maketitle
\begin{center}
    \centering
    \includegraphics[width=7.1in]{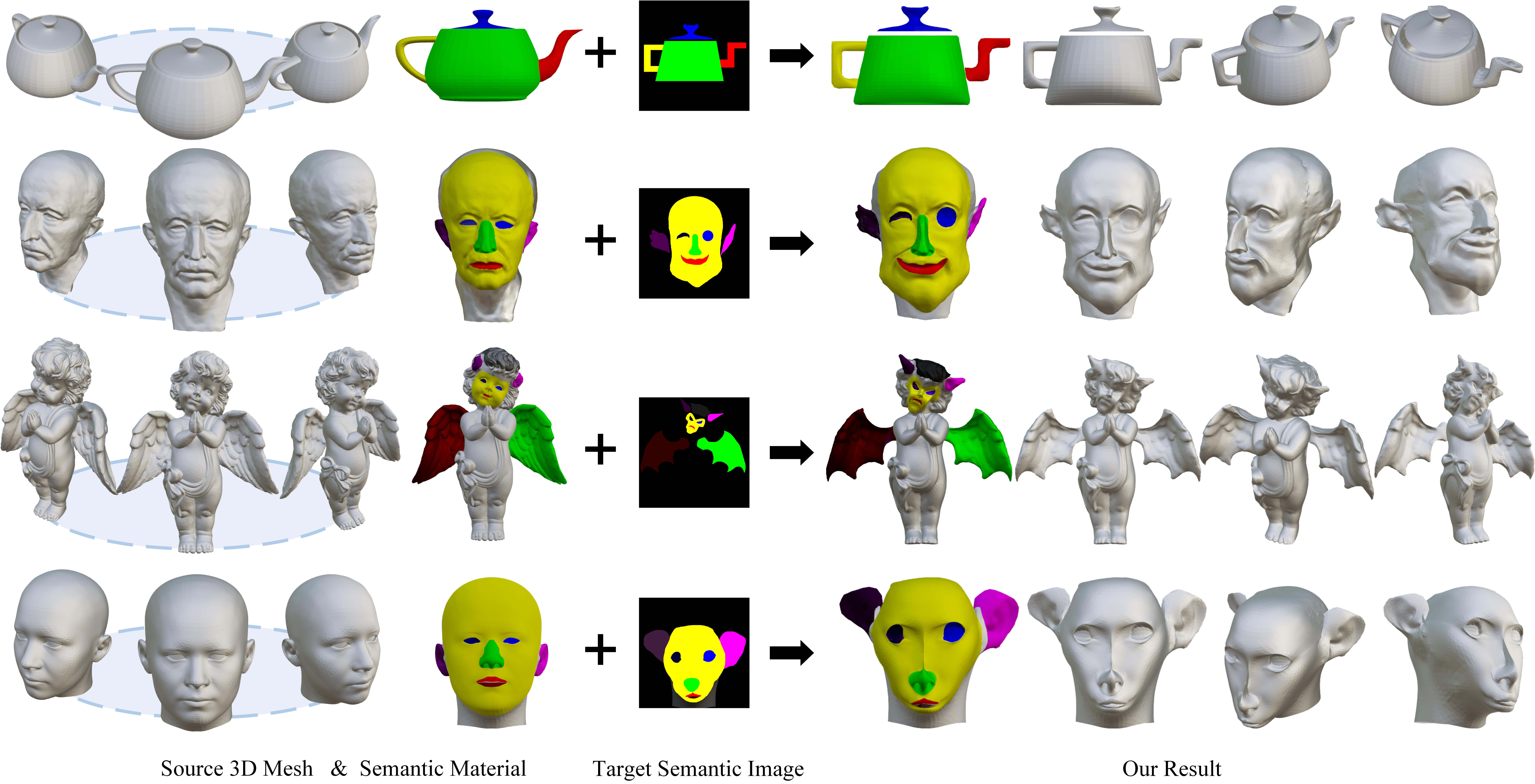}
    \captionof{figure}{3Deformer is a general-purpose framework for interactive 3D shape editing. Given a source 3D mesh with semantic materials, and a user-specified semantic image, 3Deformer can accurately edit the source mesh following the shape guidance of the semantic image, while preserving the source topology as rigid as possible. }
    \label{fig:teaser}
\end{center}%
}]

\footnotetext{
*The corresponding author, Email: niujianwei@buaa.edu.cn.

Hao Su, Jianwei Niu, Xuefeng Liu, Jiahe Cui, Ji Wan, Xinghao Wu are with State Key Lab of VR Technology and System, School of Computer Science and Engineering, Beihang University, Beijing, 100000, China. Jianwei Niu is also with Industrial Technology Research Institute, School of Information Engineering, Zhengzhou University, Henan, 450000, China; Hangzhou Innovation Institute, Beihang University, Zhejiang, 310000, China.}

\begin{abstract}
We propose 3Deformer, a general-purpose framework for interactive 3D shape editing. Given a source 3D mesh with semantic materials, and a user-specified semantic image, 3Deformer can accurately edit the source mesh following the shape guidance of the semantic image, while preserving the source topology as rigid as possible.
Recent studies of 3D shape editing mostly focus on learning neural networks to predict 3D shapes, which requires high-cost 3D training datasets and is limited to handling objects involved in the datasets.
Unlike these studies, our 3Deformer is a non-training and common framework, which only requires supervision of readily-available semantic images, and is compatible with editing various objects unlimited by datasets.
In 3Deformer, the source mesh is deformed utilizing the differentiable renderer technique, according to the correspondences between semantic images and mesh materials.
However, guiding complex 3D shapes with a simple 2D image incurs extra challenges, that is, the deform accuracy, surface smoothness, geometric rigidity, and global synchronization of the edited mesh should be guaranteed.
To address these challenges, we propose a hierarchical optimization architecture to balance the global and local shape features, and propose further various strategies and losses to improve properties of accuracy, smoothness, rigidity, and so on.
Extensive experiments show that our 3Deformer is able to produce impressive results and reaches the state-of-the-art level.
\end{abstract}



\section{Introduction}
\begin{figure*}[t]
\centering
\vspace{0cm}
\includegraphics[width=6.8 in]{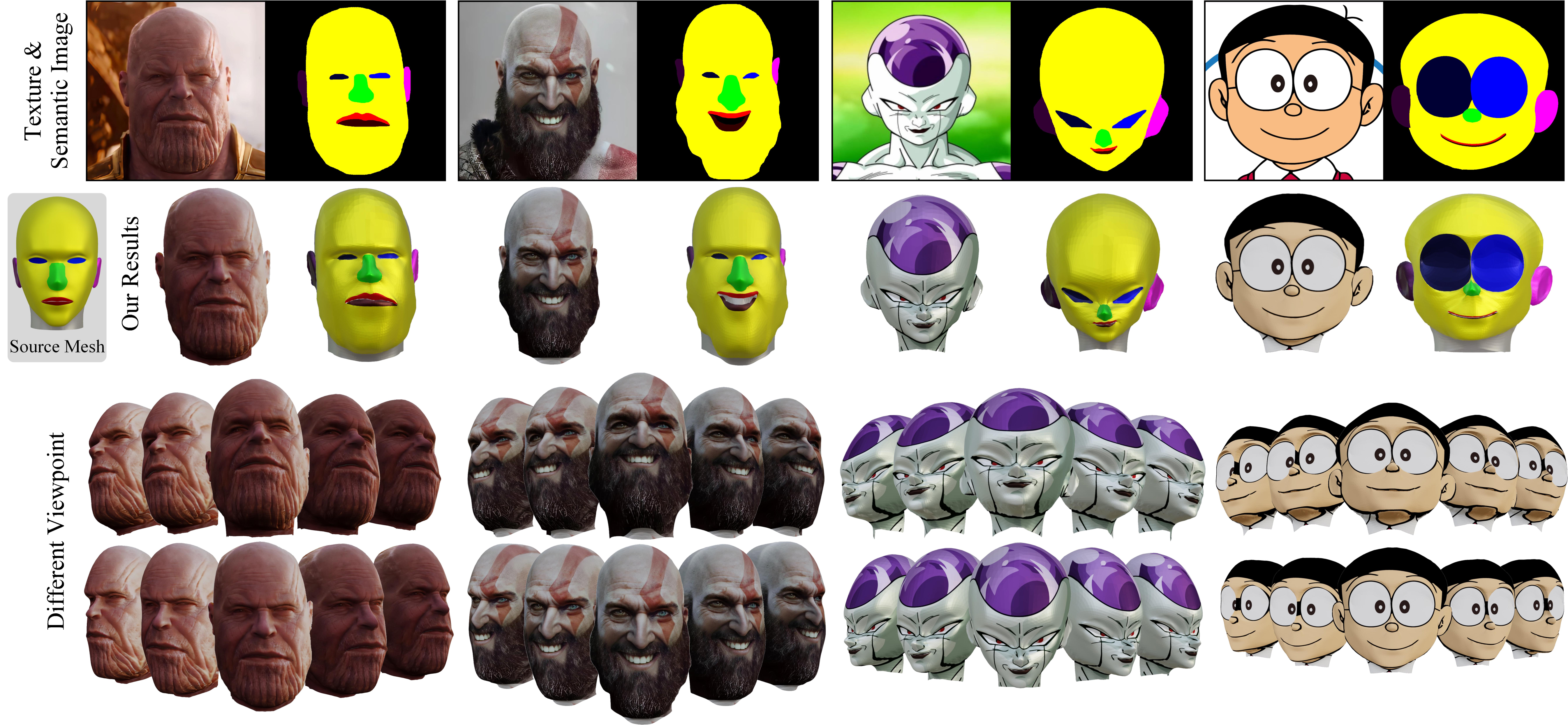}
\caption{3Deformer excels in producing 3D caricatures, which achieves accurate deformation to alignment meshes and textures, and can even produce very exaggerated facial features (e.g., cartoon, game characters). For these tasks, the learning-based methods require a large number of 3D training data in the same deformation style which are difficult to collect. Moreover, even if a model has been learned on a single deformation style, it cannot be compatible with different styles (for example, the realistic and exaggerated styles shown in the first and fourth columns).}
\label{fig:face}
\end{figure*}

3D shape editing is a longstanding task in computer graphics and vision. Existing studies have made remarkable achievements in face reconstruction (e.g., \cite{towards2020cvpr, MonocularFace2018pami, genova2018unsupervised, nicp}), 3D caricature (e.g., \cite{deepsketch2face, exemplar2021tvcg, deep2022SIGGRAPH, 3dcaricshop2021cvpr, tvcg2018caricatureshop, tog2021stylecariganv2}), shape generation (e.g., \cite{joint2021cvpr, pixel2mesh2020pami}), model editing (e.g., \cite{igarashi2005rigid, tvcg2020surface}), and so on. However, recent studies mostly focus on learning neural networks to predict edited 3D shapes by some heuristic inputs (e.g, sketches \cite{deepsketch2face}, face or object images \cite{tvcg2018caricatureshop, pixel2mesh2020pami}, point cloud \cite{3dn,fan2017point}), which requires high-cost 3D training datasets and is limited to handling object categories involved in the datasets.
Especially, as shown in Figure \ref{fig:face}, for generating 3D caricatures, cartoons, or game characters with exaggerated facial features, the learning-based methods require a large number of 3D training data in the same deformation style which are difficult to collect. Moreover, even if a model has been learned on a single deformation style, it cannot be compatible with different styles (for example, the realistic and exaggerated styles shown in the first and fourth columns of Figure \ref{fig:face}).

In this paper, we propose 3Deformer, a general-purpose and non-training framework for interactive 3D shape editing. Given a source 3D mesh with semantic materials, and a user-specified semantic image, 3Deformer can edit the source mesh to satisfy the shape guidance of the semantic image, while preserving the source geometric topology as rigid as possible.
Compared with the learning-based methods, 3Deformer has three advantages as follows. First, 3Deformer is a non-training framework that only requires the supervision of readily-available semantic images.
Second, as shown in Figure \ref{fig:teaser}, 3Deformer is not limited by training data and is applicable to editing general categories of objects (e.g., human faces, animals, and geometric solids).
Third, as shown in Figure \ref{fig:face}, 3Deformer excels in producing 3D caricatures, which achieves accurate deformation to alignment meshes and textures, and can even produce extremely exaggerated facial features (e.g., cartoons, game characters, aliens).

In 3Deformer, according to the correspondences between semantic images and source mesh materials, the mesh shapes are optimized by the technique of differentiable renderer \cite{kato2018neural, softrender, niemeyer2020differentiable,tulsiani2017multi}.
However, guiding complex 3D shapes with a simple 2D image incurs extra challenges. Specifically, we should guarantee the deform accuracy, surface smoothness, geometric rigidity, and global synchronization of the edited meshes.
To address these challenges, we propose a hierarchical optimization architecture to balance the global and local shape features, and further propose various strategies and losses to improve properties of accuracy, smoothness, rigidity, and so on.
Extensive quantitative and qualitative experiments demonstrate the effectiveness and superiority of our method which can produce impressive results and reaches the state-of-the-art level.

To summarize, our main contributions are three-fold:

\begin{itemize}\setlength{\itemsep}{5pt}
	\item We propose 3Deformer, a common non-training framework for interactive 3D shape editing, which is available for editing various object classes unlimited by training datasets.

	\item We propose a new hierarchical optimization architecture to balance the global and local shape features, and propose a binary IoU loss, a global synchronization loss, and an angle smoothing loss to guarantee the deform accuracy, surface smoothness, geometric rigidity, and global synchronization of the edited meshes.

    \item Experiments demonstrate that 3Deformer can produce impressive results for 3D shape editing, and has high performance in editing various categories of 3D objects.
\end{itemize}

\begin{figure*}[t]
\centering
\includegraphics[width=7 in]{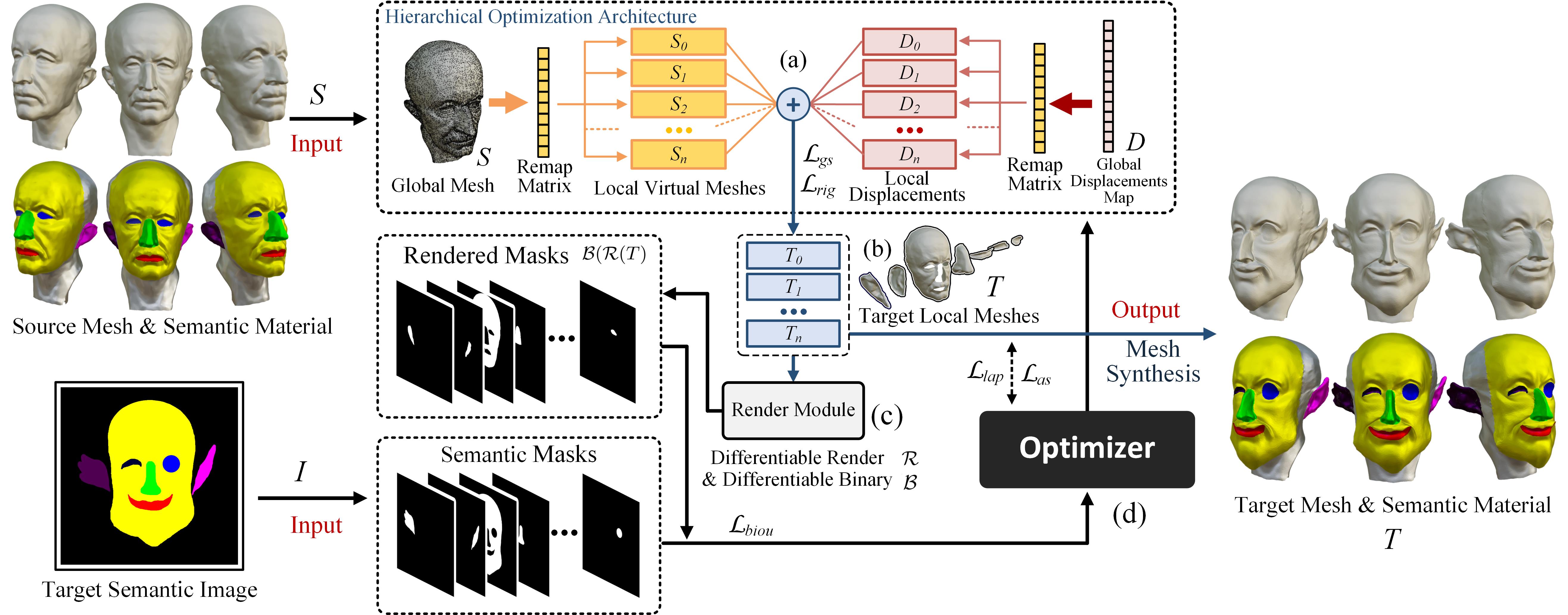}
\caption{System pipeline. Given a source mesh $S$ with semantic materials and a semantic image $I$, our method is modeled as a function $\Psi$ to deform $S$ to a target 3D mesh $T=\Psi(S, I)$. Our goal is to predict an optimal $D$ that can deform $T$ to satisfy the shape guidance of $I$, while preserving the source geometric topology as rigid as possible.}
\label{fig:method}
\end{figure*}

\section{Related Work}
Below we summarize the most related studies, involving 3D shape deformation and single-view 3D reconstruction.
\subsection{3D shape deformation}
3D mesh deformation have been actively studied in the past. Based on preserving local Laplacian properties \cite{sorkine2004laplacian} or some global features \cite{gal2009iwires}, a series of interactive editing systems has been proposed. With the progress of 2D image datasets and RGBD scans, some studies employ a reference target to guide the mesh deformation. Given source and target pairs, these studies utilize interactive \cite{TOG2011photo} or heavy processing pipelines \cite{huang2015single} to guide the deformation.
Recently, the success of deep learning has inspired learning-based studies for processing 3D models. DefFlow3D \cite{yumer2016learning} introduces a CNN architecture that produces a deformation field based on a high-level editing intent, which relies on semantic controllers and model editing results. DeformNet \cite{kurenkov2018deformnet} leverages a free-form deformation module as a differentiable layer of the network. Yet, the method produces a point set rather than a deformed mesh, and the deformation space is unsmooth. Groueix et al. \cite{eccv20183d} propose 3DCoded, a method to compute correspondences across deformable models. 3DCoded adopts an intermediate common template representation which is difficult to obtain for man-made objects. Image2mesh \cite{pontes2018image2mesh} and LFFD \cite{jack2018learning} introduce frameworks to learning the free-form deformations. Foldingnet \cite{yang2018foldingnet} deforms a 2D grid into a 3D point cloud while preserving locality features. Wang et al. \cite{3dn} propose a mesh sampling operator to train a network that deforms a source model to resemble the target, by inferring per-vertex displacements while keeping the mesh connectivity of the source model fixed. 

\subsection{Single-view 3D reconstruction}

Reconstructing 3D geometry from a single image is an ill-posed issue. Early studies adopt perceptual cues (e.g., shading \cite{horn1970shape}, texture \cite{witkin1981recovering}) to obtain a clue about surface orientations.
With the emergence of large-scale 3D datasets \cite{chang2015shapenet}, data-driven methods are presented to predict category-specific shapes directly from image features \cite{chen2019learning, choy20163d, fan2017point, girdhar2016learning, huang2015single, mescheder2019occupancy, pontes2018image2mesh, wang2018pixel2mesh}, including voxel \cite{choy20163d, girdhar2016learning}, point cloud \cite{fan2017point}, triangle mesh \cite{huang2015single, pontes2018image2mesh, wang2018pixel2mesh}, and implicit function \cite{chen2019learning, mescheder2019occupancy}. Sketch2Model \cite{sketch2model} introduces a neural framework to produce 3D meshes from single free-hand sketches.
Moreover, differentiable renderer (DR) techniques \cite{kato2018neural, softrender, niemeyer2020differentiable,tulsiani2017multi} are proposed to preserve the differentiability of rendering meshes into images, which enables the image-wise supervision to backpropagate to the 3D meshes.

For the reviewed studies, compared with the learning-based methods that require high-cost 3D training datasets, our 3Deformer only requires supervision of a readily-available semantic image, which is general-purpose and compatible with editing various categories of objects. Compared with the DR-based methods, we propose a series of improvements to overcome the challenge that guiding complex 3D shapes with a simple single-view image, which improves the deform accuracy, surface smoothness, geometric rigidity, and global synchronization of the edited meshes.

\section{Method}
\subsection{Overview}
 Given a source mesh $S$ with semantic materials and a user-specified semantic image $I$, our method is modeled as a function $\Psi$ to deform $S$ to a target mesh $T$$=$$\Psi(S, I)$.
 We represent the source mesh as $S$$=$$(V,E$$)$, where $V$$\in$$\mathbb{R}^{N_v\times3}$ is the $(x, y, z)$ coordinates of vertices, and $E\in \mathbb{Z}^{N_e\times3}$ is the edges of triangle surfaces. $N_v$ and $N_e$ are the numbers of vertices and triangles respectively. $T$ is computed by $S$ and the per-vertex displacements $D\in \mathbb{R}^{N_v\times3}$, formulated as $T$$=$$(V$$+$$D,E)$.

Figure \ref{fig:method} shows our overall pipeline.
First, as shown in Figure \ref{fig:method}(a), leveraging a designed hierarchical optimization architecture (HOA), we remap global vertices (or global displacements) of $S$ (or $T$) into a local mesh sequence $\Phi_s=\{S_n|S_n$$\in$$S\}$ (or local displacement sequence $\Phi_d$$=$$\{D_n|D_n$$\in$$D\}$), according to the correspondences between $S$ and $I$.
Then, as shown in Figure \ref{fig:method}(b), in each iteration, we optimize $\Phi_d$ to minimizing our loss function as Eq.(\ref{equation:ltotal}), and calculate the target mesh $\Phi_t=$$\{T_n|T_n$$\in$$T\}$ by $\Phi_s$ and $\Phi_d$.
Simultaneously, as shown in Figure \ref{fig:method}(c), $T_n$ is rendered to a 2D mask $\mathcal{B}(\mathcal{R}(T_n))$ by differentiable rendering $\mathcal{R}$ \cite{softrender} and differentiable binarizing $\mathcal{B}$, and $\mathcal{B}(\mathcal{R}(T_n))$ is encouraged to be visually similar to mask $I_n$, where $I_n$ is separated from $I$ following different user-specified colors.

To sum up, our goal is to predict an optimal $D$ that can accurately deform $T$ to satisfy the shape supervision of $I$, while preserving the source topology as rigid as possible. $D$ is optimized by minimizing our designed loss function $\mathcal{L}_{total}$, represented as

\begin{equation}
\begin{aligned}
\mathcal{L}_{total} & =  \lambda_1\mathcal{L}_{biou}(T,I)+ \lambda_2 \mathcal{L}_{gs}(S,T) + \\
& \lambda_3 \mathcal{L}_{ag}(T) + \lambda_4 \mathcal{L}_{rig}(S,T)  + \lambda_5 \mathcal{L}_{lap}(T)
\end{aligned},
\label{equation:ltotal}
\end{equation}
where $\lambda_1$ to $\lambda_5$ are used to balance the multiple objectives. Each term of $\mathcal{L}_{total}$ is detailed in Section \ref{sec:loss}.


\begin{figure}[t]
\centering
\includegraphics[width=3.3 in]{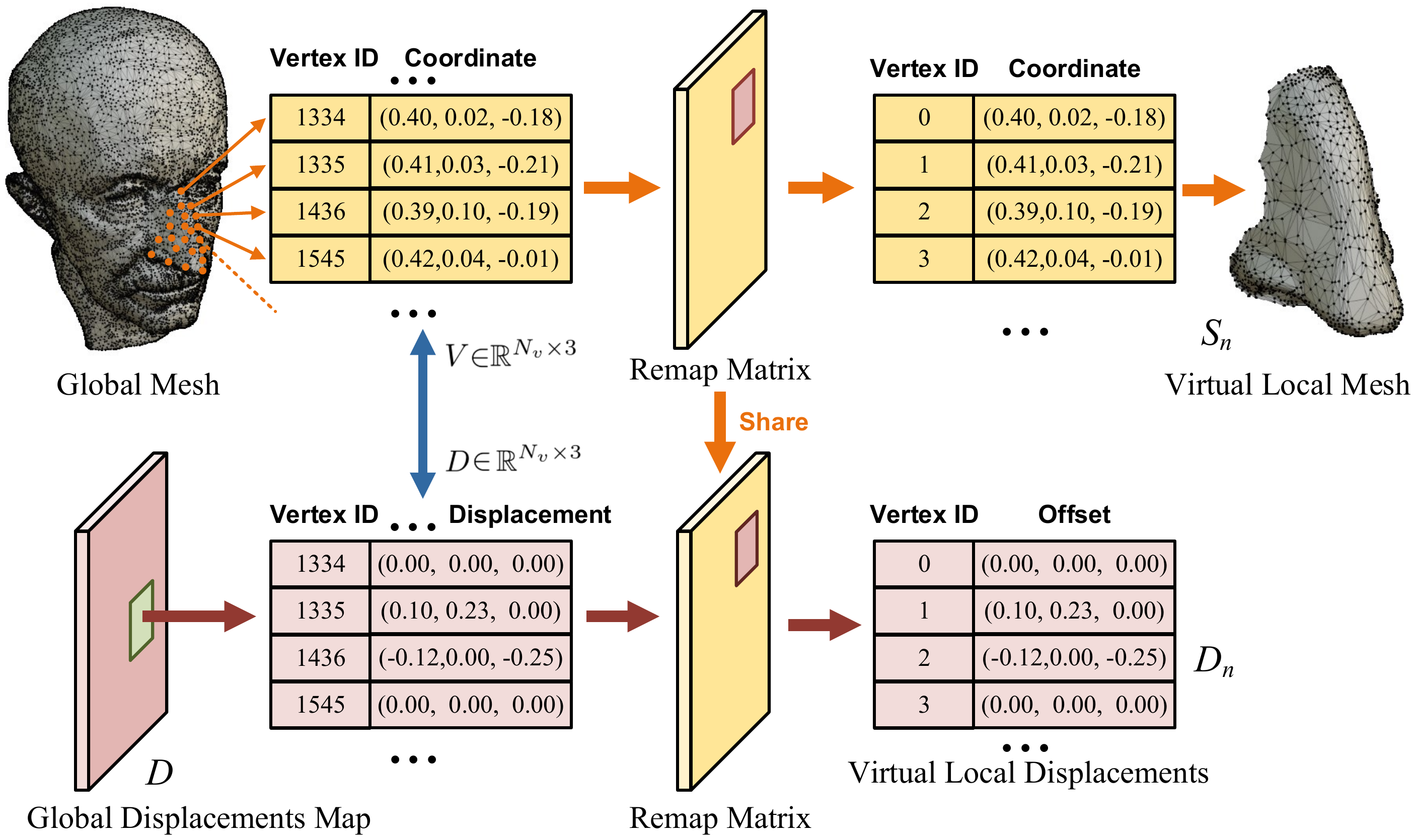}
\caption{Framework of the hierarchical optimization architecture (HOA). Instead of copying geometric data (e.g., vertices, displacements) in $S$ (or $D$) to construct a new local mesh (or displacements), we create a remap matric to map the data in $S$ (or $D$) to build virtual local meshes $S_n$ (or displacements $D_n$).}
\label{fig:hoa}
\end{figure}

\subsection{{Hierarchical optimization architecture}}
As shown in Figure \ref{fig:method}(a)-(b), the hierarchical optimization architecture (HOA) is designed to remap $S$ and $D$ to local meshes $\Phi_s$ and local displacements $\Phi_d$ respectively. %

The underlying idea of HOA is shown in Figure \ref{fig:hoa}. Specifically, instead of copying geometric data (e.g., vertices, edges, triangles, and displacements) of $S$ (or $D$) to construct a new local mesh (or displacements), we create a remap matric to map the data of $S$ (or $D$) to build a virtual local mesh $S_n$ (or displacement $D_n$). In other words, each local mesh (or displacement) is only a remapping of the global mesh (or global displacement), and thus the local and global features can be optimized and balanced synchronously.

The advantages of HOA are two-fold.
First, the deformation of each local mesh is directly mirrored to the global mesh, which enables to control and balance the global and local features (e.g., shape, smoothness, and rigidity).
Second, it preserves the differentiability of global-local conversion, and guarantees $D$ can be controlled by the loss back-propagation.

\begin{figure}[t]
\centering
\includegraphics[width=3.3 in]{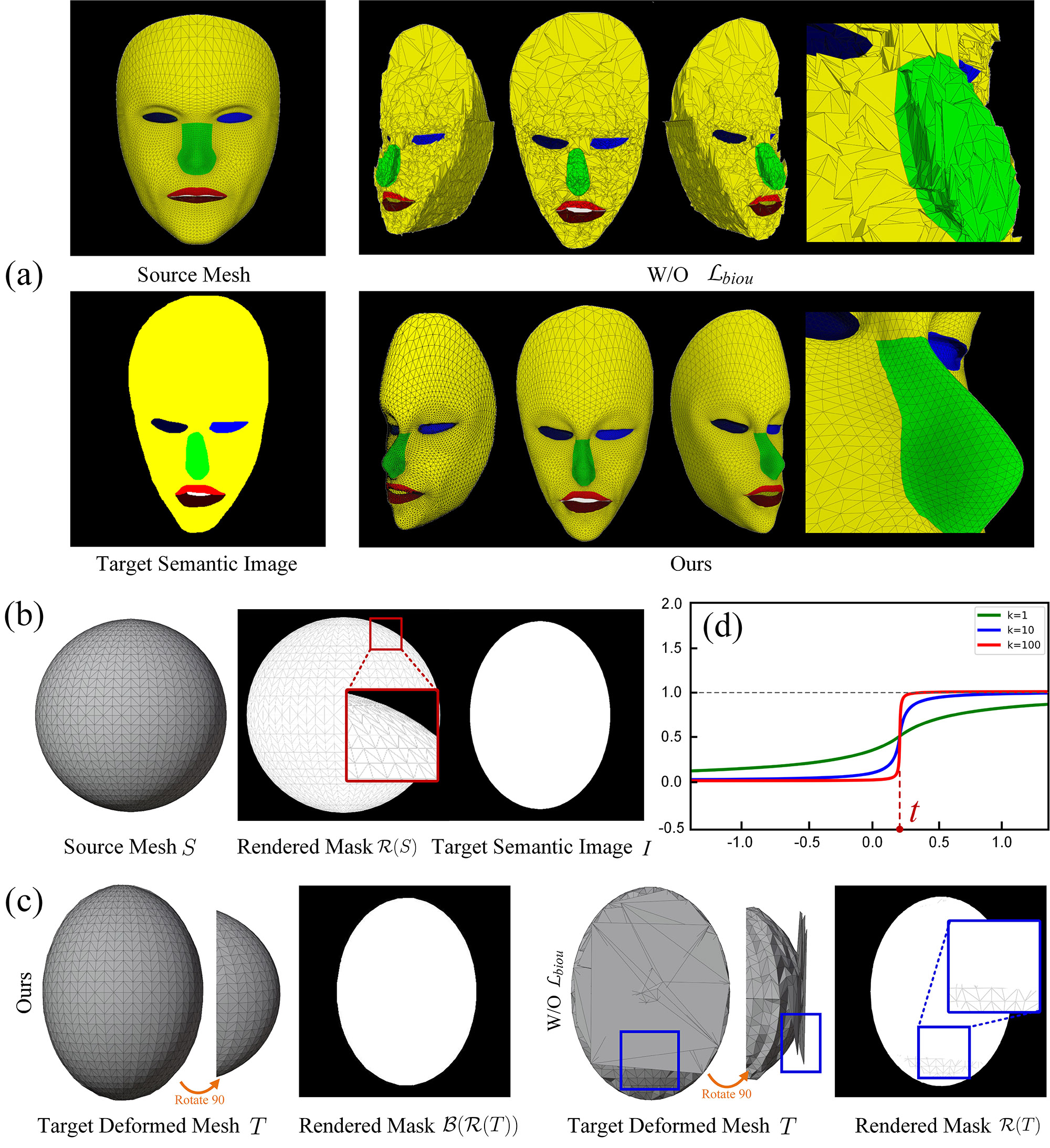}
\caption{(a) $\mathcal{L}_{biou}$ effectively addresses the issue of rugged and irregular surfaces.
(b)(c) Without $\mathcal{L}_{biou}$, $\mathcal{R}$ will rend edges and triangle surfaces in different colors (i.e., gray and white, marked by red boxes), and surfaces of $T$ will be expanded or stretched to occlude edges (blue boxes) for making $\mathcal{R}(T)$ more similar to $I$. (d) Curve of the differentiable binary term $\mathcal{B}$.}
\vspace{-0.2cm}
\label{fig:biou}
\end{figure}


\subsection{Loss function}
\label{sec:loss}
Our loss function $\mathcal{L}_{total}$ consists of five terms: binary IoU loss $\mathcal{L}_{biou}$, global synchronization loss $\mathcal{L}_{gs}$, angle smooth loss $\mathcal{L}_{as}$, rigidity loss $\mathcal{L}_{rig}$, and Laplacian loss $\mathcal{L}_{lap}$, where $\mathcal{L}_{biou}$, $\mathcal{L}_{vs}$, and $\mathcal{L}_{ag}$ are our proposed terms, and $\mathcal{L}_{rig}$ \cite{amberg2007optimal} and $\mathcal{L}_{lap}$ \cite{softrender} are proposed by previous works.

\begin{figure}[t]
\centering
\includegraphics[width=3.2 in]{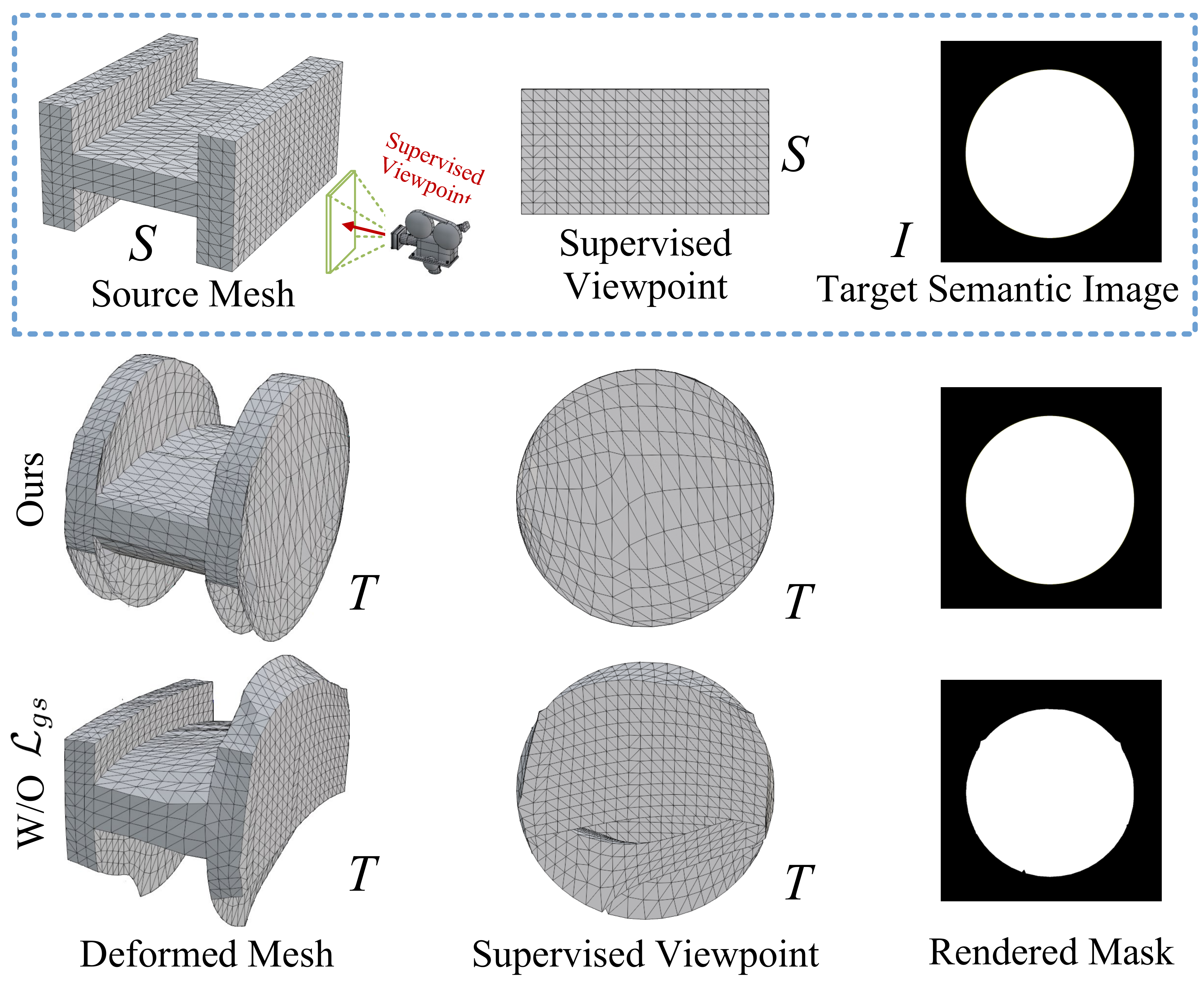}
\caption{$\mathcal{L}_{gs}$ significantly guarantees the global consistency and rigidity of deformation, including the occluded vertices from the supervised viewpoint.}
\label{fig:global_loss1}
\end{figure}

\textbf{Binary IoU loss.}
The IoU (Intersection over Union) loss \cite{kato2018neural, softrender} is typically used to minimize the error between target image $I$ and rendered mesh $\mathcal{R}(T)$. However,
as shown in Figure \ref{fig:biou}(a)(c), we notice that if only using the IoU loss, $\mathcal{R}$ will render edges and triangle surfaces in different colors (i.e., gray and white, marked by red boxes). Hence, surfaces of $T$ will be expanded or stretched to occlude edges (blue boxes) to achieve a high similarity between $\mathcal{R}(T)$ and $I$, which incurs the issue of rugged and irregular surfaces.

 To address this issue, we propose a binary IoU loss $\mathcal{L}_{biou}$ defined as
\begin{equation}
\begin{aligned}
\!\!\!\!\mathcal{L}_{biou}(I_n,\!\mathcal{R}(T_n)) & \! =\!  1\!-\!\frac{\| I_n\!\!\otimes \!\mathcal{B}(\mathcal{R}(T_n))\|_1}{\| I_n\!\! \oplus  \! \mathcal{B}(\mathcal{R}(T_n)) - I_n\!\!  \otimes \! \mathcal{B}(\mathcal{R}(T_n))\|_1}\!\!\!\!\!
\end{aligned},
\label{equation:biou}
\end{equation}
where $I_n$$\in$$I$, $T_n$$\in$$T$, and $\mathcal{B}$ is our proposed differentiable binary term, defined as
\begin{equation}
\begin{aligned}
\mathcal{B}(\mathcal{R}(T_n))  =\frac{\alpha(\mathcal{R}(T_n),t)}{1+\sqrt{\alpha(\mathcal{R}(T_n),t)^2}}
\end{aligned},
\label{equation:biou2}
\end{equation}
where $\alpha(\mathcal{R}(T_n),t) = k$$(\mathcal{R}(T_n) - t)$. Figure \ref{fig:biou}(d) shows the curve of $\mathcal{B}$, $t$ is the threshold of differentiable binarization, and $k$ is used to control the curve slope. Empirically, we set $t$$=$$0.5$ and $k$$=$$100$.

With the help of $\mathcal{L}_{biou}$, the influence of gray edges is significantly avoided without disturbing the differentiability, and our method is able to pay close attention to optimizing the surfaces' shapes, smoothness, and rigidity.

\begin{figure}[t]
\centering
\includegraphics[width=3.2 in]{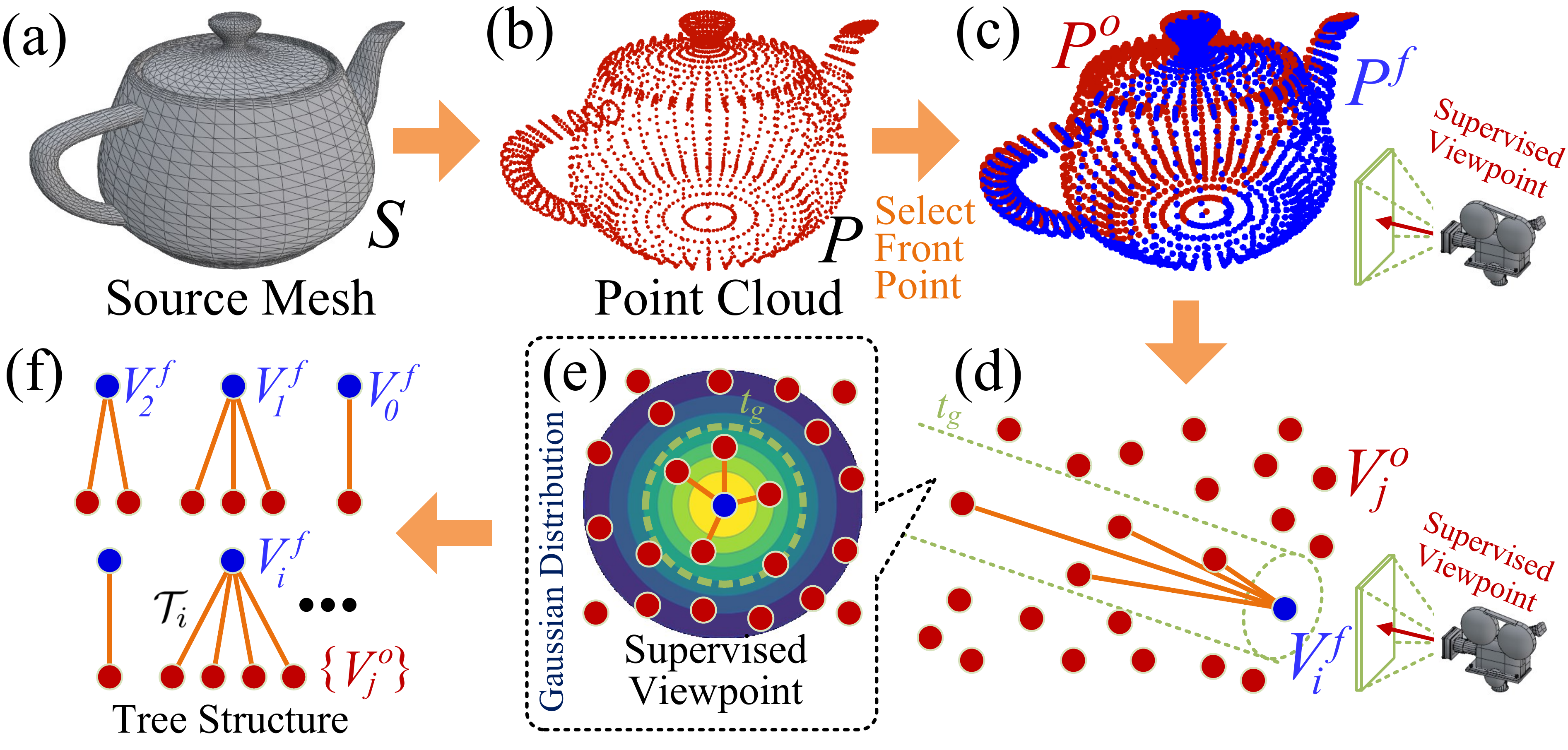}
\caption{Initialization of calculating $\mathcal{L}_{gs}$. We match front vertices $P^f$ and nearby occluded vertices $P^o$, and construct a set of trees. The main idea of $\mathcal{L}_{gs}$ is that we encourage the observable front vertices $V^f_i$ to guide the deformation of occluded vertices $V^o_j$.}
\label{fig:global_loss2}
\end{figure}

\textbf{{Global synchronization loss.}} As shown in Figure \ref{fig:global_loss1}, in single-view 3D reconstruction, the solution of deformation is not unique due to the simplicity of guidance, which results in producing undesired deformation. To address this issue, we propose a global synchronization loss $\mathcal{L}_{gs}$ to guarantee the global consistency and rigidity of deformation, including the occluded vertices from the supervised viewpoint.

Figure \ref{fig:global_loss2} shows the initialization of calculating $\mathcal{L}_{gs}$. First, we convert $S$ to a point cloud $P$, and each point in $P$ indicates a vertex of $S$ [Figure \ref{fig:global_loss2}(a)(b)]. Then, $P$ is divided into front points $P^f$ (blue) and occluded points $P^o$ (red) [Figure \ref{fig:global_loss2}(c)]. From the supervised viewpoint, for each front vertex $V^f_i$$\in$$P^f$, we calculate a Gaussian distribution $\mathcal{G}_{V^f_i}$ as
\begin{equation}
\begin{aligned}
\mathcal{G}_{V^f_i}(x,y)  = \frac{1}{2\pi \sigma^2}e^{\frac{x^2+y^2}{2 \sigma^2}}
\end{aligned},
\label{equation:gauss}
\end{equation}
where $(x, y)$ are coordinates on the projection plane with origin at $V^f_i$.
Next, we match $V^f_i$ and occluded vertices $V^o_j$$\in$$P^o$ whose coordinates are $(x_j,y_j)$, and the calculated weights $\mathcal{G}_{V^f_i}(x_j,y_j)$ are greater than threshold $t_g$ [Figure \ref{fig:global_loss2}(d)(e)].
Finally, for each $V^f_i$, we construct a tree structure $\mathcal{T}_i$ which sets $V^f_i$ as a root note and all matched vertices in set $\{V^o_j | V^o_j$$\in$$P^o$$,\mathcal{G}_{V^f_i}(x_j,y_j)$$>$$t_g\}$ as leaf notes [Figure \ref{fig:global_loss2}(f)]. Based on the above initialization, $\mathcal{L}_{gs}$ is defined as
\begin{equation}
\begin{aligned}
\!\!\!\!\mathcal{L}_{gs}(D) &  =  \sum_{V^f_i \! \in P^f}\sum_{V^o_j  \in \mathcal{T}_i}\| (D_{V^f_i} - D_{V^o_j})\cdot\mathcal{G}_{V^f_i}(x_j,y_j) \|^2_2
\end{aligned},
\label{equation:gs}
\end{equation}
where $D_{V^f_i}$ and  $D_{V^o_j}$ are displacements of vertices $V^f_i$ and ${V^o_j}$ respectively.

The underlying idea of $\mathcal{L}_{gs}$ is that for the matched root-leaf vertices in each tree, we try to make them have similar displacements after global deformation. In other words, $\mathcal{L}_{gs}$ aims to guide the deformation of occluded vertices by the observable front vertices.

\begin{figure}[t]
\centering
\includegraphics[width=3.2 in]{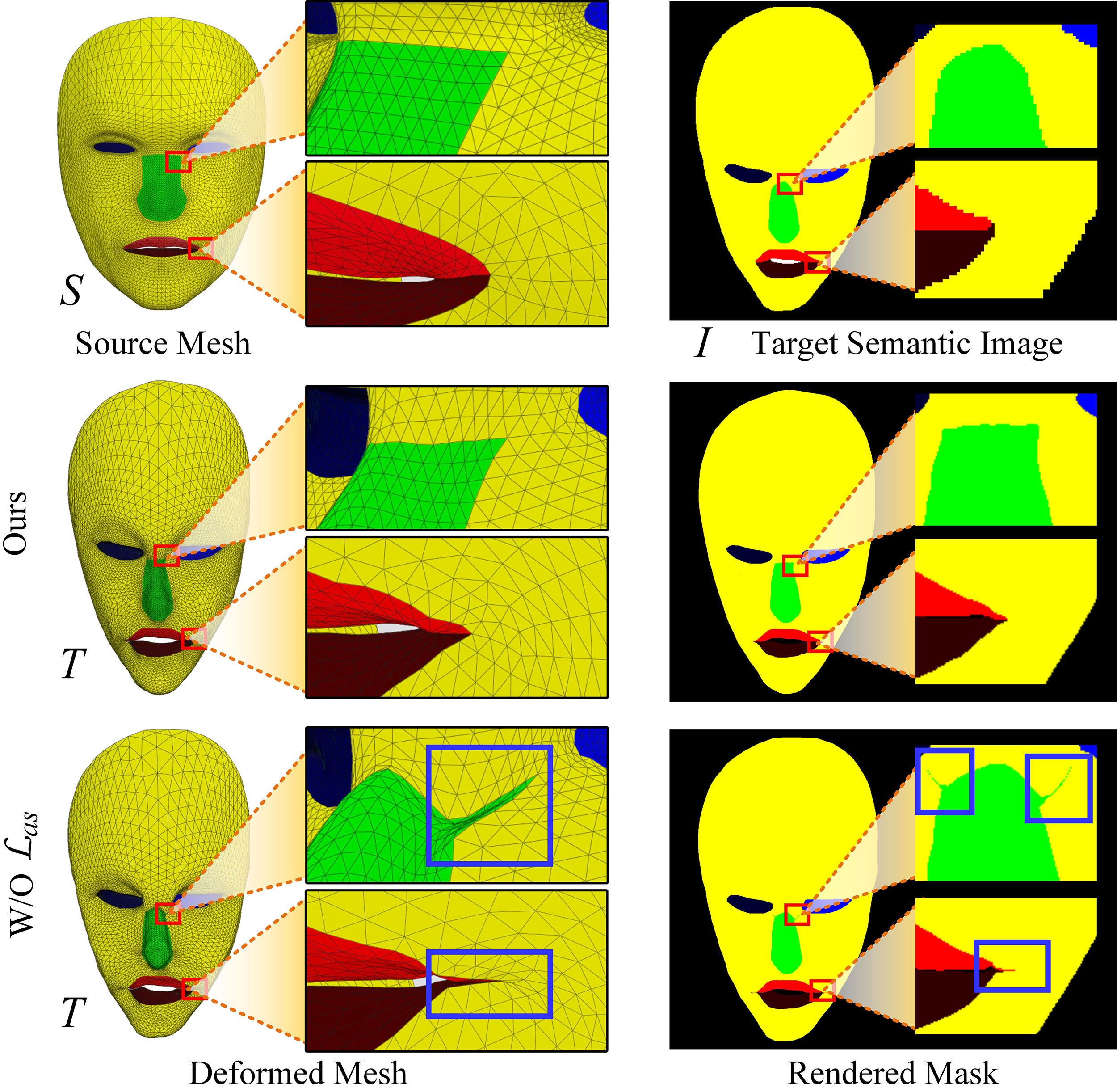}
\caption{$\mathcal{L}_{as}$ addresses the issue of {surfaces flattening}. Aiming at decreasing the error between $I$ and $\mathcal{B}(\mathcal{R}(T))$, some triangle surfaces of $T$ will be flattened into undesired lines (blue boxes).}
\label{fig:angle_loss}
\end{figure}

\begin{figure}[t]
\centering
\includegraphics[width=3.2 in]{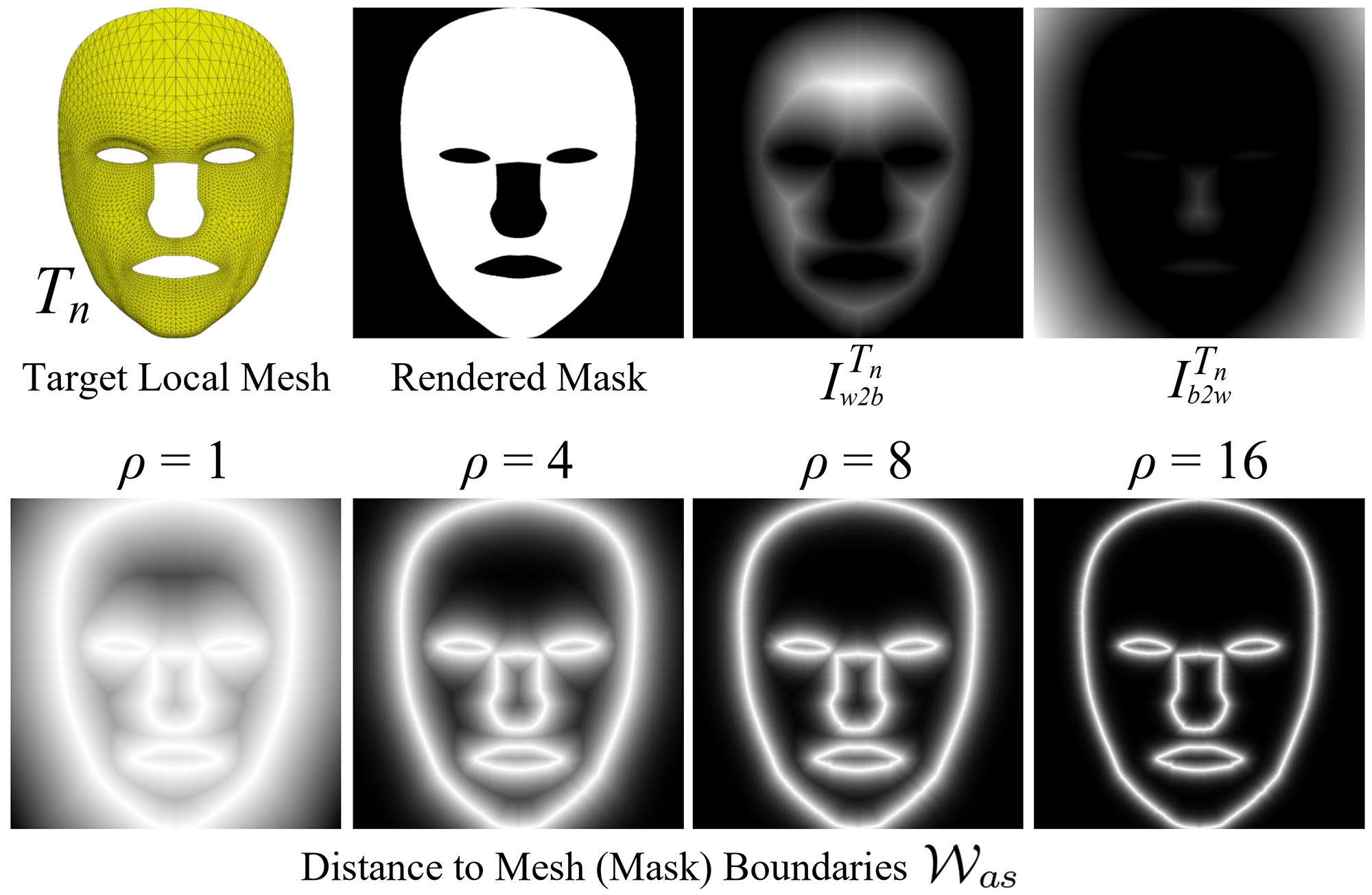}
\caption{Definition of the boundary distance weight $\mathcal{W}_{as}$ that stores the distance of each pixel to its nearest mask (mesh) boundary.  }
\label{fig:angle_loss2}
\end{figure}

\begin{figure*}[!ht]
\centering
\vspace{-2cm}
\includegraphics[width=6.8 in]{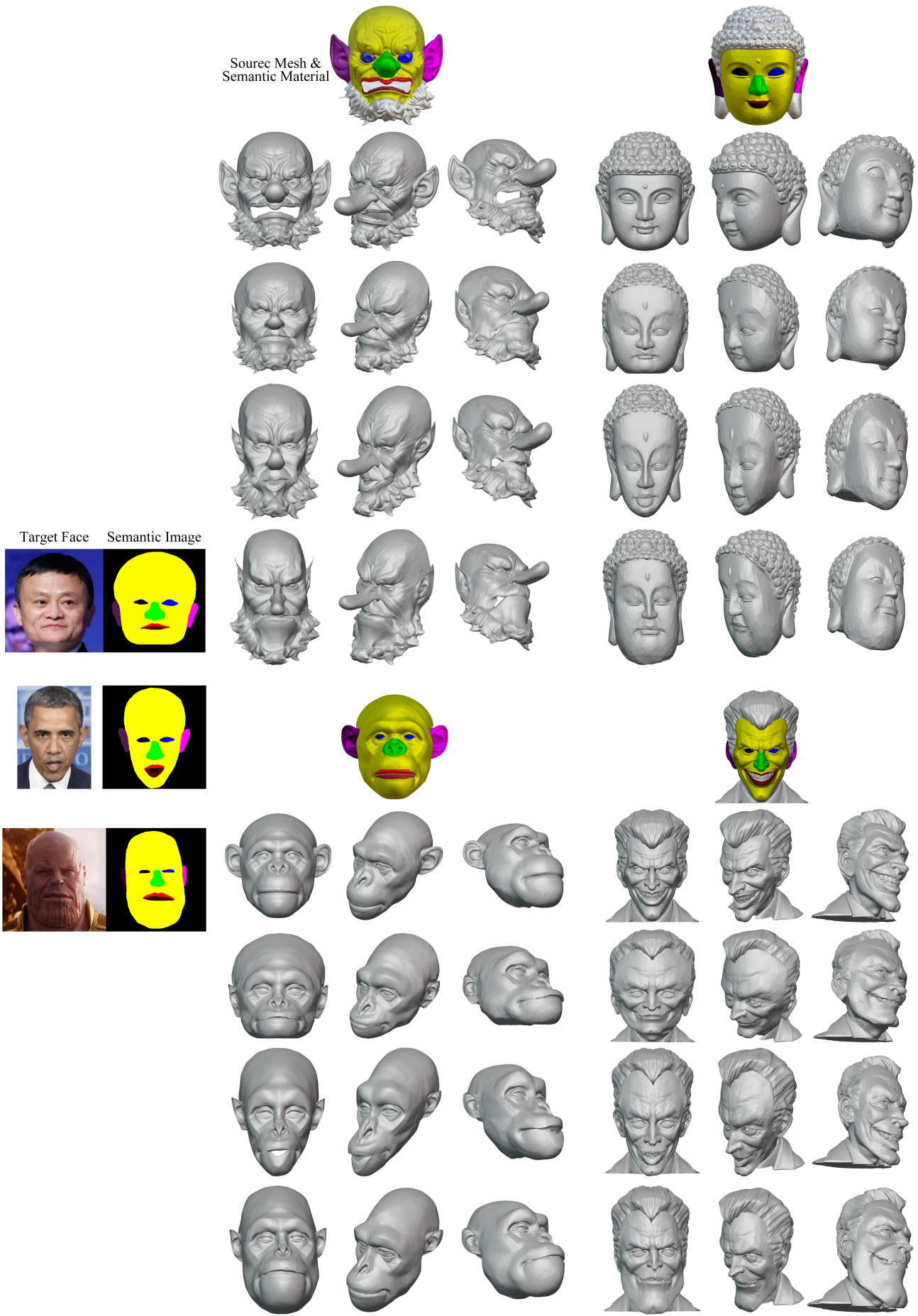}
\caption{Visual comparison of editing different source meshes by different semantic images. 3Deformer is able to edit source meshes with extremely complex shapes, while preserving source topologies as rigid as possible.}
\label{fig:sample}
\end{figure*}

\textbf{Angle smooth loss.} As shown in Figure \ref{fig:angle_loss}, in deformed mesh $T$, some triangle surfaces will be flattened into undesired lines (blue boxes). Although the flattened surfaces decrease the error between $I$ and $\mathcal{B}(\mathcal{R}(T))$, they compromise our performance in preserving geometric topology and shape smoothness.
To overcome this defect, we propose an angle smooth loss $\mathcal{L}_{as}$.

The underlying idea of $\mathcal{L}_{as}$ is that we encourage a triangle's angle closer to mesh boundaries to be greater. $\mathcal{L}_{as}$ is designed in an image-graphic synergistic manner. As shown in Figure \ref{fig:angle_loss2}, first, for a local mesh $T_n$, we render a binary mask $\mathcal{R}(\mathcal{B}(T_n))$, and compute two distance images $I^{T_n}_{w2b}$ and $I^{T_n}_{b2w}$.
 Each pixel in $I^{T_n}_{w2b}$ (or $I^{T_n}_{b2w}$) stores the distance value to its nearest black (or white) pixel. Then, a boundary distance weight $\mathcal{W}_{as}$ is defined as
\begin{equation}
\begin{aligned}
\mathcal{W}_{as} = [\frac{1}{2}(I^{T_n}_{b2w} + I^{T_n}_{w2b})]^\rho
\end{aligned},
\label{equation:Was}
\end{equation}
where $\mathcal{W}_{as}$ stores the distance of each pixel to its nearest mask (mesh) boundary, $\mathcal{W}_{as}\in [0, 1]$, and $\rho$ balances the attention level of $\mathcal{W}_{as}$.
Next, we map $\mathcal{W}_{as}$ to the 3D graphic space of $T_n$, and endow each vertex $v_i\in T_n$ with a weight value $\mathcal{W}_{as}(v_i)$, and $\mathcal{L}_{as}$ is defined as
\begin{equation}
\begin{aligned}
\!\!\!\!\mathcal{L}_{as}(T_n) &  = \sum_{\theta_j\in T_n} (1 + \mathrm{cos} (\theta_j)) \cdot \mathcal{W}_{as}(v_i)
\end{aligned},
\label{equation:angle}
\end{equation}
where $\theta_j$ is a angle corresponding to vertex $v_i$ in a triangle surface, $\theta_j \in T_n$, and $\mathcal{W}_{as}(v_i)$ is inversely proportional to the distance between $v_i$ and its nearest mesh boundary.

\section{Experiment}
Below we evaluate our performance in 4 aspects: visual perception, editing accuracy, ablation study, and time cost.
\subsection{Implementation}
We implement our program in PyTorch and all experiments are performed on a computer with an NVIDIA Geforce RTX 2080 GPU, and 16 Intel(R) i7-10700F CPU.
For all experiments, by default, we set $\lambda_1=10^9$, $\lambda_2=10^3$, $\lambda_3=10^3$, $\lambda_4=10^6$, and $\lambda_5=10^2$ in eq.(\ref{equation:ltotal}), $\rho=16$ in eq.(\ref{equation:Was}). The resolutions of input semantic images $I$ are $512\times512$, the learning rate is $10^{-3}$, and $T$ is output after $2$$\times$$10^3$ iterations.

\subsection{Evaluation of editing different meshes}
We evaluate our performance for editing different meshes in subjective visual perception and objective deformation accuracy.

\textbf{Visual perception.} Figure \ref{fig:teaser}, Figure \ref{fig:face}, and Figure \ref{fig:sample} compare our results of editing different source meshes by different target images. Experimental results demonstrate that our method can significantly edit source meshes with extremely complex shapes, while preserving geometric topology as rigid as possible.

\begin{figure}[ht]
\centering
\includegraphics[width=3.2in]{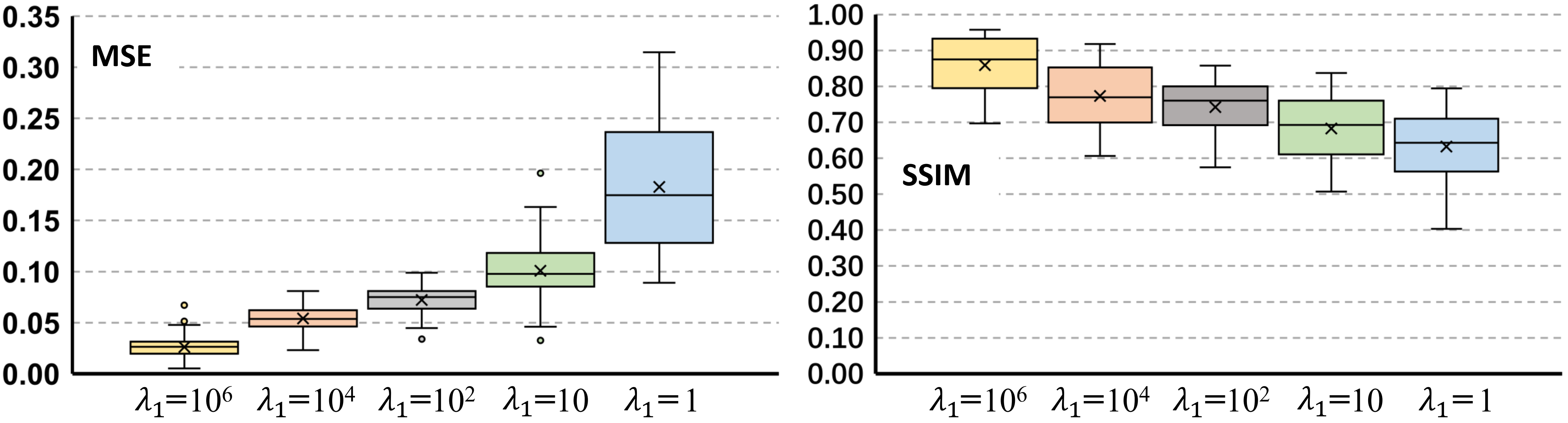}
\caption{Comparison of deformation accuracies under different settings of $\lambda_1$ in Eq.(\ref{equation:ltotal}).}
\label{fig:accuracy}
\end{figure}

\textbf{Accuracy.} To evaluate our edited accuracies objectively, we adopt indexes MSE (Mean Square Error) and SSIM \cite{ssim} (Structural SIMilarity) to measure the similarities between rendered mesh $\mathcal{B}(\mathcal{R}(T))$ and target semantic image $I$. MSE index is computed by $\frac{\|\mathcal{B}(\mathcal{R}(T)) - I\|^2_2}{CHW \times 255}$$\in$$[0,1]$. SSIM index is in [-1, 1], and the index value is proportional to the similarity.

Since $\mathcal{L}_{biou}$ controls the visual similarities between $\mathcal{B}(\mathcal{R}(T))$ and $I$, we compare the editing accuracies by progressively modifing the weight $\lambda_1$ and fixing the other weights in Eq.(\ref{equation:ltotal}). In each setting of $\lambda_1$, we produce 50 results guided by 5 source meshes and 10 semantic images, and the summarized results are shown in Figure \ref{fig:accuracy}. Experimental results show that our generated meshes are accurately similar to the guided image when $\lambda_1\geqslant 10^6$, and $\mathcal{L}_{biou}$ is effective to control the editing accuracy.

\begin{figure}[t]
\centering
\includegraphics[width=3.3 in]{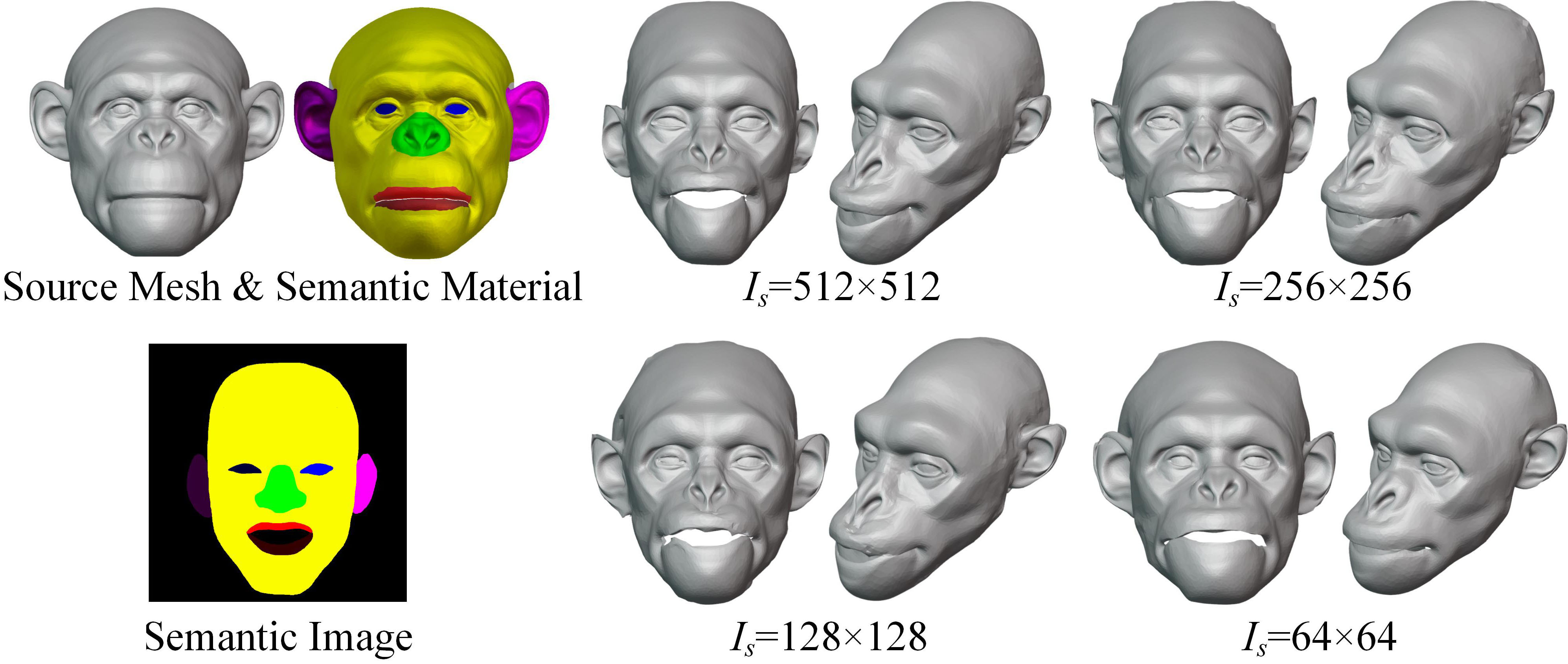}
\caption{Influences of semantic $\!$image $\!$resolution $\!I_s\!\!$ .}
\label{fig:diff_size}
\end{figure}

\textbf{Influence of semantic image resolution.} As shown in Figure \ref{fig:diff_size}, we evaluate the influences of semantic image resolution $I_s$ on mesh editing. When $I_s$ are low (e.g., $I_s$$\leqslant$$128$$\times$$128$), the mesh shapes cannot be clearly guided and result in producing undesired results. Empirically, we set $I_s$ to $512$$\times$$512$ or $1024$$\times$$1024$. In Sec. \ref{sec:time}, we evaluate our time costs under different settings of $I_s$.

\begin{figure}[t]
\centering
\includegraphics[width=3.5in]{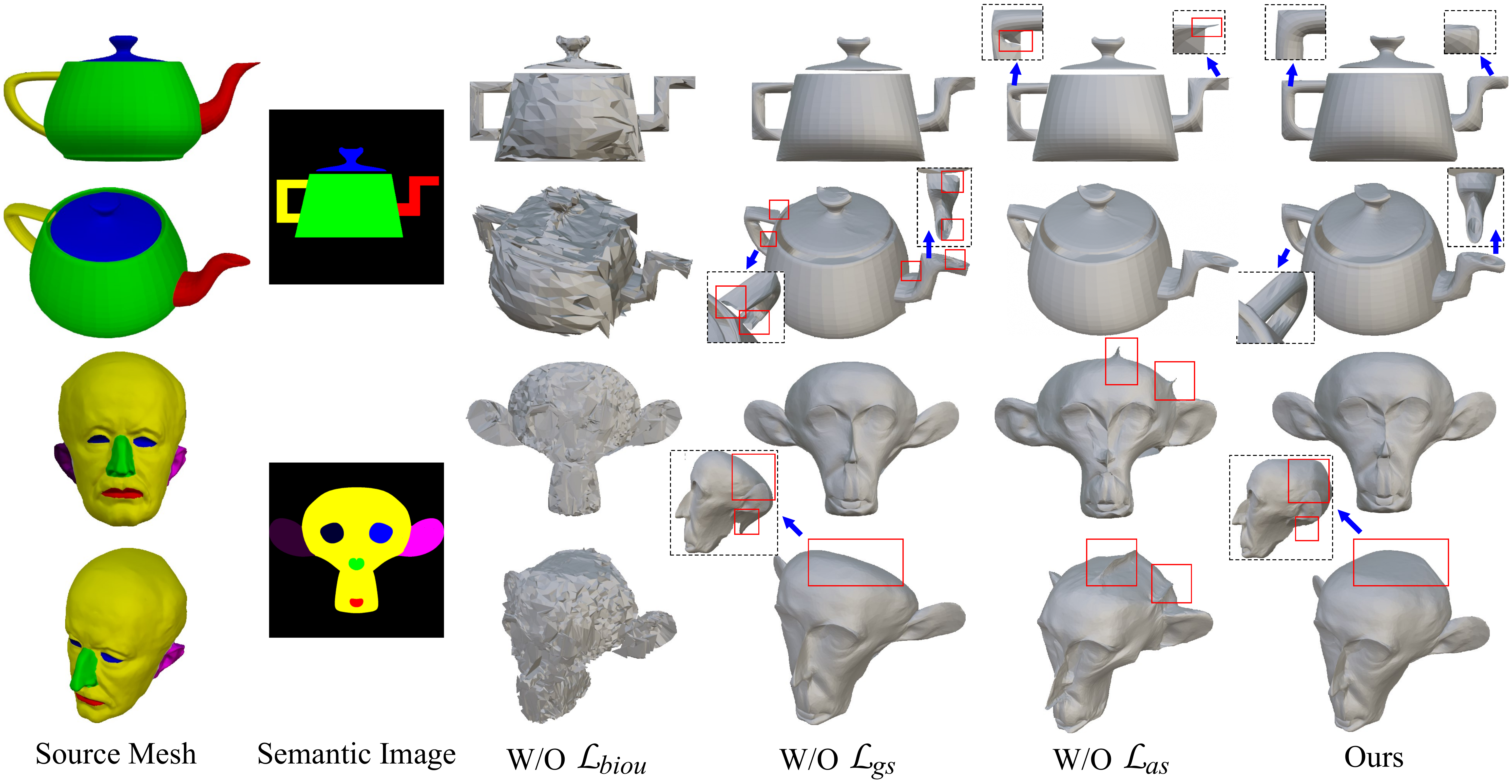}
\caption{Ablation study on our improvements. The results show that each of our improvements is essential to produce high-quality results.}
\label{fig:ablation_study}
\end{figure}

\begin{figure*}[t]
\centering
\includegraphics[width=7 in]{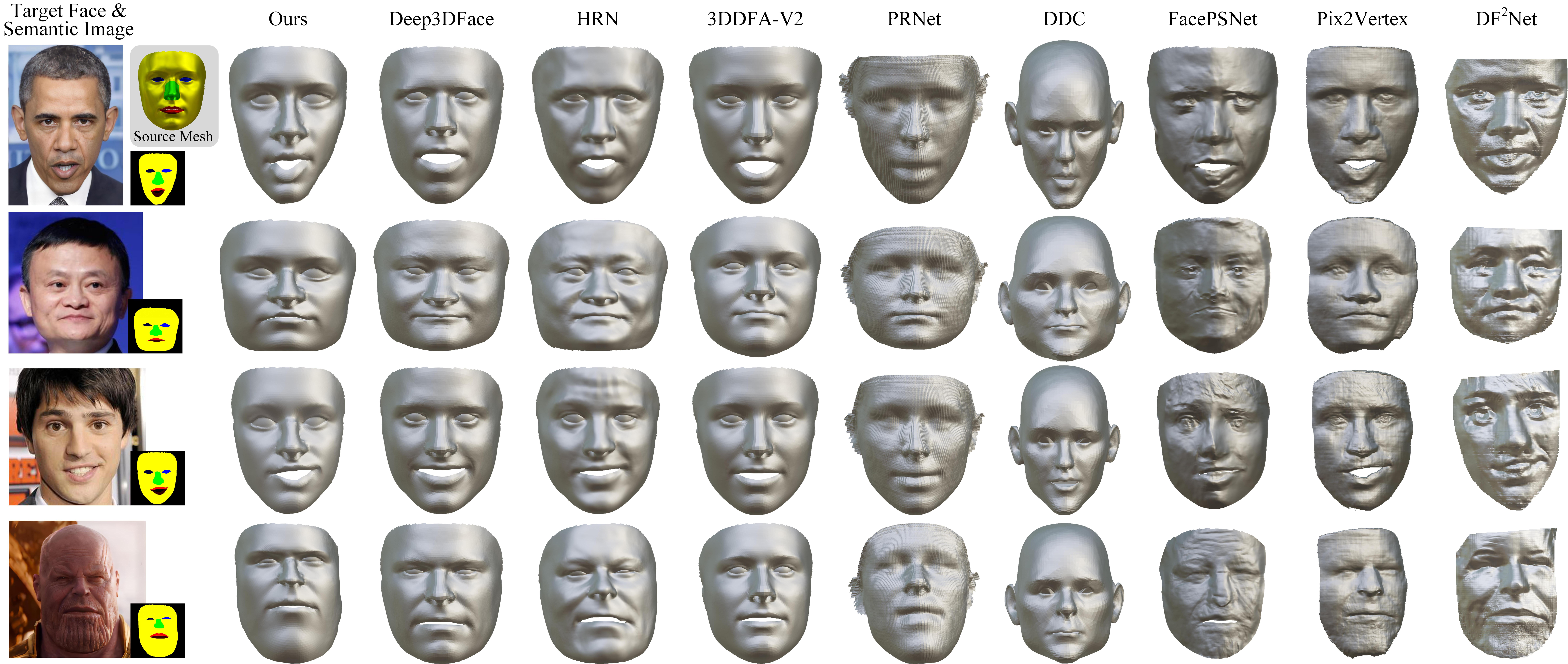}
\caption{Compared with other methods of face reconstruction and 3D caricature, including Deep3DFace \cite{deng2019accurate}, HRN \cite{HRN}, 3DDFA-V2 \cite{towards}, PRNet \cite{prn}, DDC \cite{jung2022deepdeformable}, FacePSNet \cite{Lightweight}, Pix2Vertex \cite{sela2017unrestricted}, and DF$^2$Net \cite{DF2Net}. Without using training data, our method can produce results similar to other state-of-the-art methods. }
\label{fig:face_reconstruction}
\end{figure*}

\subsection{Ablation study}
Below we conduct an ablation study to evaluate each of our improvement, and the generated results under different improvements are shown in Figure \ref{fig:ablation_study}.

The binary IoU loss $\mathcal{L}_{biou}$ is essential to avoid the issue of rugged and irregular surfaces. Without $\mathcal{L}_{biou}$ (replaced by IoU loss), surfaces of deformed meshes will be irregularly warped [Figure \ref{fig:biou}(a)].
The global synchronization loss $\mathcal{L}_{gs}$ is essential to guarantee the global consistency of deformation. Without $\mathcal{L}_{gs}$, the deformation of occluded vertices cannot be supervised, which reduces the performance of preserving the global rigidity (red boxes) (Figure \ref{fig:global_loss2}).
The angle smooth loss $\mathcal{L}_{as}$ is essential to resist the surface flattening issue. Without $\mathcal{L}_{as}$, some triangle surfaces in $T$ will be flattened into undesired lines (red boxes) (Figure \ref{fig:angle_loss2}). The ablation study shows that each of our improvements is effective and essential to produce high-quality results.

\subsection{Comparison with other methods}
\textbf{3D face recognition.} We compare our method with state-of-the-art methods of 3D face recognition and 3D caricature, including Deep3DFace \cite{deng2019accurate}, HRN \cite{HRN}, 3DDFA-V2 \cite{towards}, PRNet \cite{prn}, DDC (Deep Deformable 3D Caricatures) \cite{jung2022deepdeformable}, FacePSNet \cite{Lightweight}, Pix2Vertex \cite{sela2017unrestricted}, and DF$^2$Net \cite{DF2Net}.
As shown in Figure \ref{fig:face_reconstruction}, without using any training data, our method can produce results similar to other state-of-the-art methods. 

\begin{figure}[t]
\centering
\includegraphics[width=3.4 in]{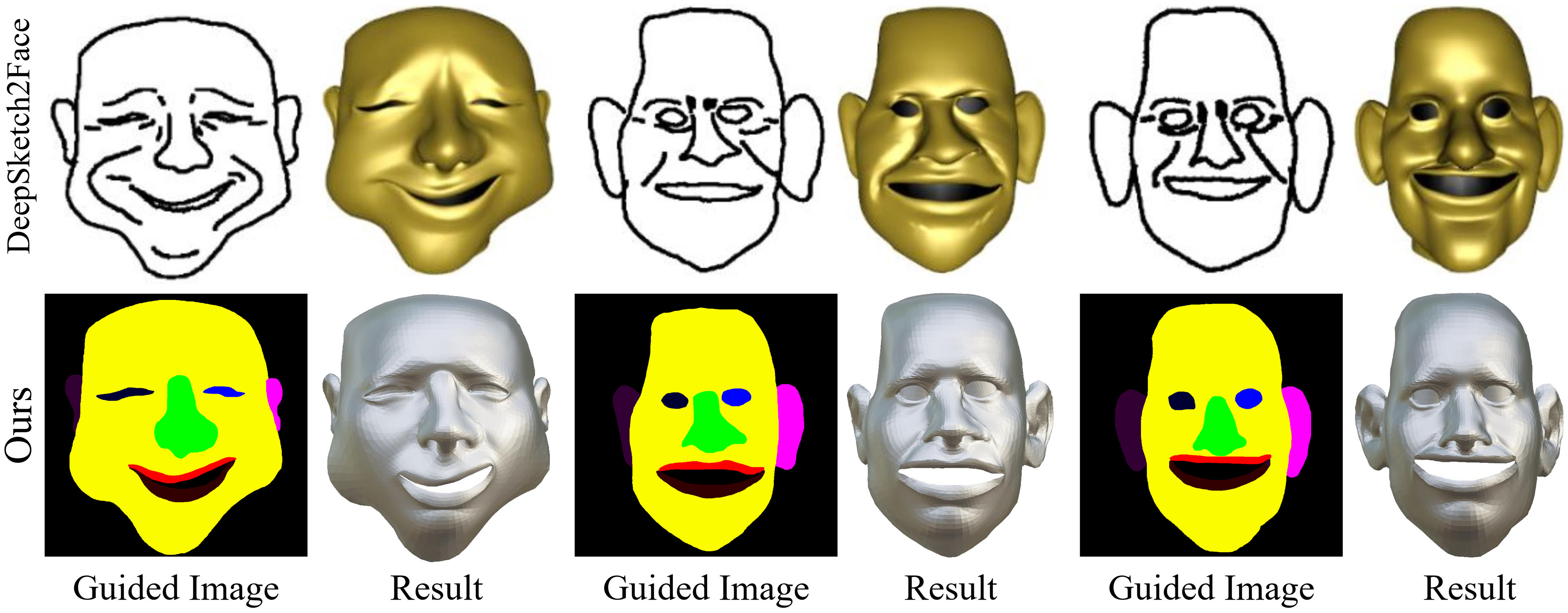}
\caption{Compared with the image-guided method DeepSketch2Face \cite{deepsketch2face} (our source mesh as shown in the last row of Figure \ref{fig:teaser}). Our 3Deformer can produce similar results without any training data.}
\label{fig:image_guided}
\end{figure}

\textbf{Image-guided caricature.} We also compare our method with DeepSketch2Face \cite{deepsketch2face} that predicts 3D faced by inputting face sketches. Comparison results in Figure \ref{fig:image_guided} show that our method can produce similar results with DeepSketch2face, without using any training data. Moreover, as shown in Figure \ref{fig:teaser}, Figure \ref{fig:face}, and Figure \ref{fig:accuracy}, our method is not limited by the training datasets and is available to various object classes, and can produce extremely exaggerated faces with different deformation styles (e.g., cartoons, game characters, aliens).

\subsection{Time cost}
\label{sec:time}
 We evaluate our time costs under different resolutions $I_s$ of semantic images and different numbers of edited vertices $N_v$. Each result is output in 2000 iterations.

 The summarized results in Table \ref{tab:compare} show that our time cost is proportional to $N_v$, and less influenced by $I_s$. In addition, our method takes an acceptable 750-800 seconds to edit a mesh with approximately 15000 vertices, which is an acceptable time cost in practical applications.

\renewcommand\arraystretch{1.3}
\begin{table}[t]\small
\centering
\caption{Time costs (seconds) under different settings of $I_s$ and $N_v$.}
\begin{tabular}{c|m{1cm}<{\centering}|m{0.8cm}<{\centering}|m{0.8cm}<{\centering}|m{0.8cm}<{\centering}|m{0.8cm}<{\centering}}
\hline
\hline
\diagbox[width=1.6cm, height=0.8cm]{$I_s\!\!\!\!$}{$N_v (\approx)\!\!\!$} & {20000} &{15000} &{10000} &  {5000}& {1000} \\
\hline

$56\times56$   & 928.26  & 763.65 & 618.54 & 462.63 &346.82\\
\hline
$128\times128$ & 921.65  & 769.02 & 613.07 & 471.83 & 355.61  \\
\hline
$256\times256$ & 933.94  & 772.65 & 596.83 & 468.07 & 352.71 \\
\hline
$512\times512$ & 918.37  & 756.34 & 605.40 & 474.12 & 349.07 \\
\hline
\hline
\end{tabular}
\label{tab:compare}
\end{table}
\renewcommand\arraystretch{1}

\subsection{Application and limitation}
\label{application}
\textbf{Shape editing.} Figure \ref{fig:teaser} and Figure \ref{fig:sample} show that without any training data, our 3Deformer effectively edits the mesh shapes while preserving geometric topology and as rigid as possible. Furthermore, 3Deformer is general-purpose and applicable to editing a wide range of object categories (e.g., human faces, animals, and geometric solids).

\textbf{3D caricature.} As shown in Figure \ref{fig:face} and Figure \ref{fig:image_guided}, 3Deformer is effectively applied to generate 3D caricatures, which achieves accurate deformation to align meshes and textures, and even can produce a face with extremely exaggerated facial features (e.g., cartoon, alien, and game characters).

\textbf{Build dataset.} 2D semantic images are readily generated or modified (e.g., translating, scaling, deforming) by a data augmentation program or manual operation. For each modified semantic image, 3Deformer can easily produce a corresponding 3D mesh, which is applicable to building a dataset for training an end-to-end neural model that predicts 3D meshes from 2D semantic images.

\textbf{Limitation.} First, by default, our 3Deformer can only control the shape from one supervised viewpoint (e.g., front or customized camera position). If requiring guidance from additional viewpoints, more semantic images and camera positions should be provided. Second, compared with learning-based methods, 3Deformer is slower in running time. If using 3Deformer to build a dataset of paired semantic images and deformed meshes, a neural model may be trained to fast produce exaggerated 3D shapes from 2D semantic images. We leave those for future work.

\section{Conclusion}
In this paper, we propose a non-training framework for interactive 3D shape editing, named 3Deformer, which only requires the guidance of a readily-available semantic image without high-cost 3D datasets.
3Deformer is general-purpose and compatible with editing a wide range of object categories, and can produce high-quality 3D shapes with preserving the source geometric topologies as rigid as possible.
Extensive experiments show that 3Deformer can produce impressive results and reaches the state-of-the-art level.

\bibliographystyle{ieee_fullname}
\bibliography{ref}

\begin{IEEEbiography}[{\includegraphics[width=1in]{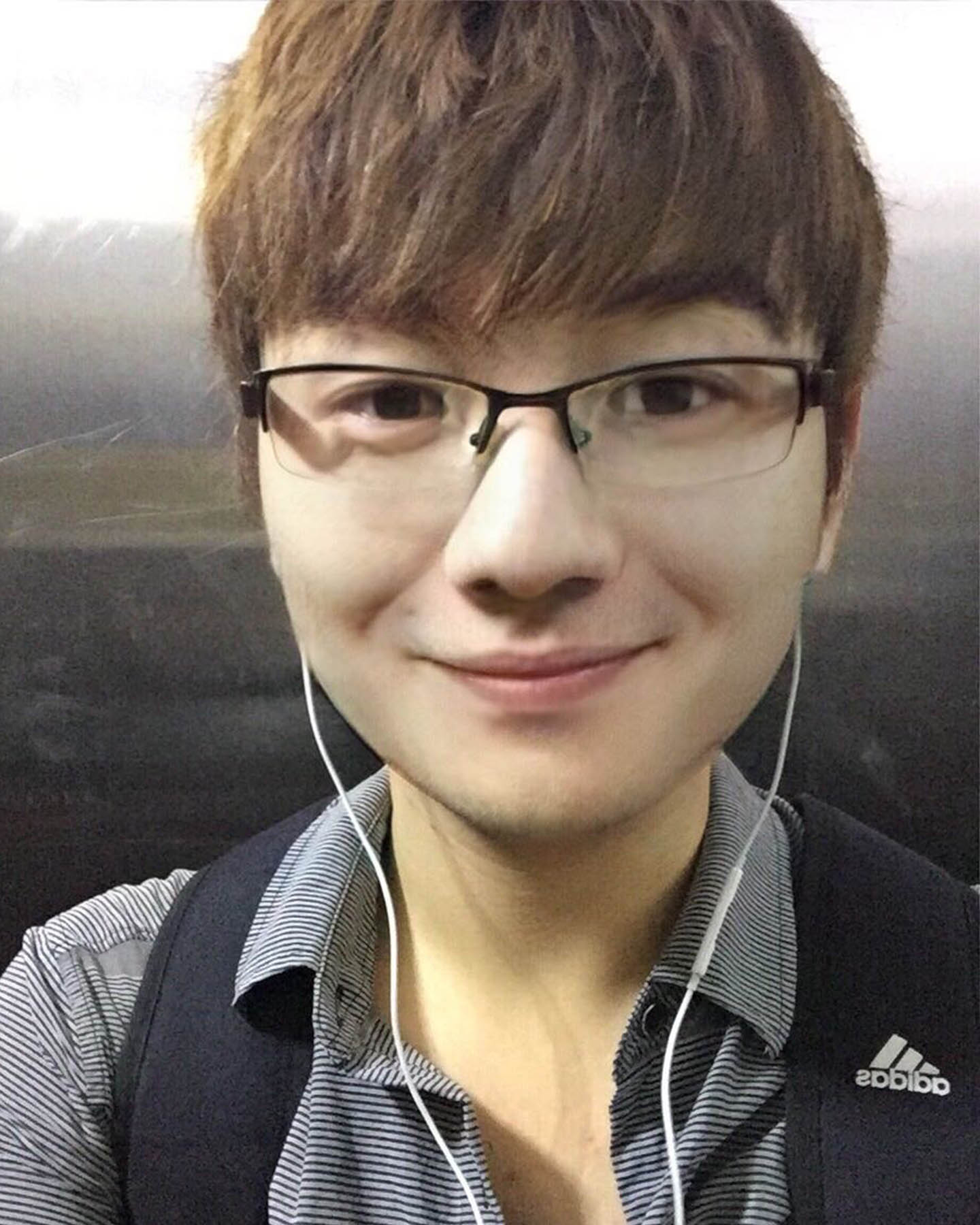}}]{Hao Su}
is a Ph.D. student in State Key Laboratory of Virtual Reality Technology and Systems, Beihang University, China. He received the B.E. degree in Computer Science and Technology from Zhengzhou University, in 2016. His research interests are in deep learning, computer graphics, and image processing.
\end{IEEEbiography}

\begin{IEEEbiography}[{\includegraphics[width=1in,height=1.25in,clip,keepaspectratio]{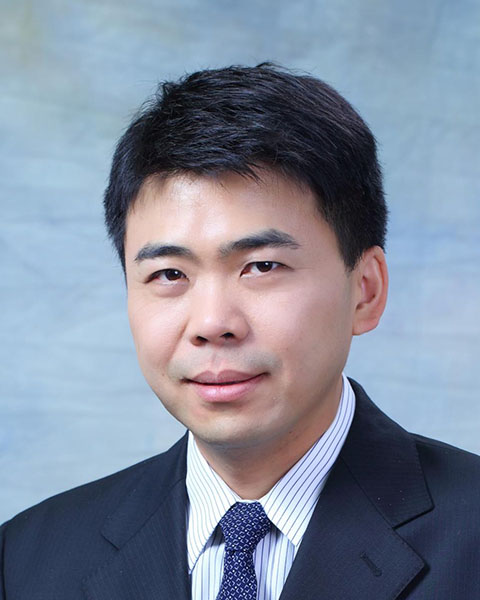}}]{Xuefeng Liu}
 received the M.S. and Ph.D. degrees from the Beijing Institute of Technology, China, and the University of Bristol, United Kingdom, in 2003 and 2008, respectively. He was an associate professor at the School of Electronics and Information Engineering in the HuaZhong University of Science and Technology, China from 2008 to 2018. He is currently an associate professor at the School of Computer Science and Engineering, Beihang University, China. His research interests include wireless sensor networks, distributed computing and in-network processing. He has served as a reviewer for several international journals/conference proceedings.
\end{IEEEbiography}

\begin{IEEEbiography}[{\includegraphics[width=1in,height=1.25in,clip,keepaspectratio]{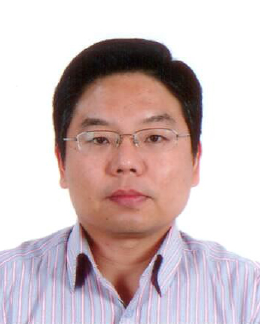}}]{Jianwei Niu}
 received the M.S. and Ph.D. degrees in computer science from Beihang University, Beijing, China, in 1998 and 2002, respectively. He was a visiting scholar at School of Computer Science, Carnegie Mellon University, USA from Jan. 2010 to Feb. 2011. He is a professor in the School of Computer Science and Engineering, BUAA, and an IEEE senior member. His current research interests include mobile and pervasive computing, mobile video analysis.
\end{IEEEbiography}

\begin{IEEEbiography}[{\includegraphics[width=1in,height=1.25in,clip,keepaspectratio]{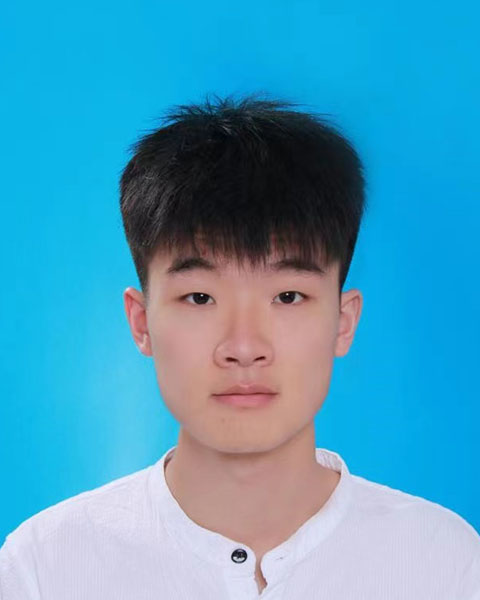}}]{Jiahe Cui}
is currently working toward the PhD degree in college of computer and science at Beihang University, Beijing, China. His research interests are multi-sensor SLAM and perceptual algorithms in autonomous driving.
\end{IEEEbiography}

\begin{IEEEbiography}[{\includegraphics[width=1in,height=1.25in,clip,keepaspectratio]{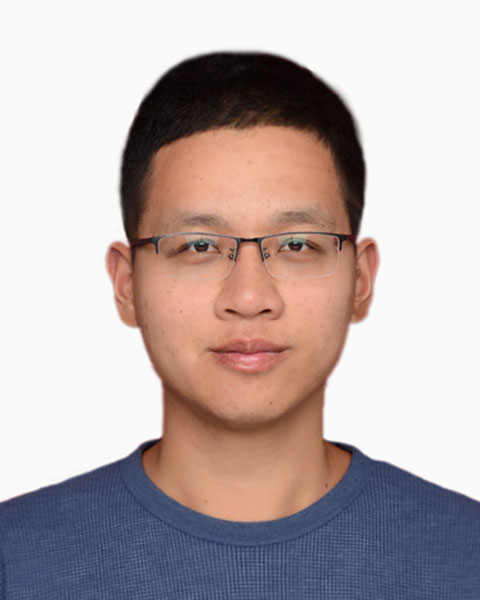}}]{Ji Wan}
is currently working toward the PhD degree at the State Key Laboratory of Software Development Environment, Beihang University. His research interests include distributed systems and blockchain.
\end{IEEEbiography}

\begin{IEEEbiography}[{\includegraphics[width=1in,height=1.25in,clip,keepaspectratio]{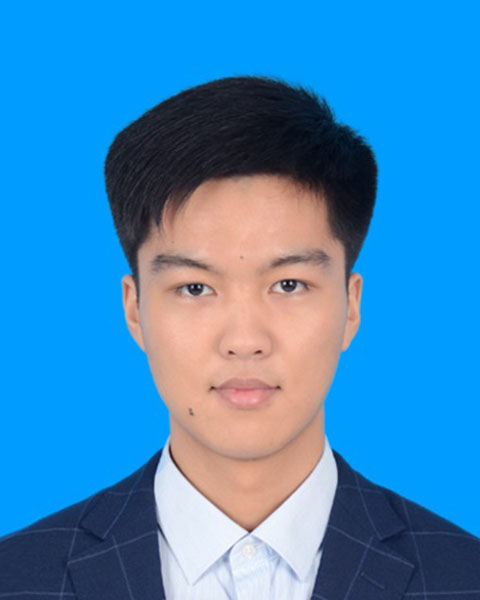}}]{Xinghao Wu}
is currently working toward the PhD degree in the school of computer science and engineering, Beihang University. His research interests are federated learning and distributed machine learning.
\end{IEEEbiography}
\end{document}